\pdfoutput=1
\documentclass{article}

\usepackage[preprint, nonatbib]{neurips_2023}

\usepackage[utf8]{inputenc} 
\usepackage[T1]{fontenc}    
\usepackage{hyperref}       
\usepackage{url}            
\usepackage{booktabs}       
\usepackage{amsfonts}       
\usepackage{nicefrac}       
\usepackage{microtype}      
\usepackage{xcolor}         
\usepackage{graphicx}
\usepackage{amsmath, bm}
\usepackage[numbers]{natbib}

\title{Contextual Feature Extraction Hierarchies Converge in Large Language Models and the Brain}

\author{%
  Gavin Mischler\thanks{These authors contributed equally to this work} \\
  Department of Electrical Engineering\\
  Columbia University\\
  New York, NY 10027 \\
  \texttt{gm2944@columbia.edu} \\
  \And
  Yinghao Aaron Li\textsuperscript{$\ast$} \\
  Department of Electrical Engineering\\
  Columbia University\\
  New York, NY 10027 \\
  \texttt{yl4579@columbia.edu}\\
  \And
  Stephan Bickel \\
  The Feinstein Institutes for Medical Research\\
  Northwell Health\\
  Manhasset, NY 11030 \\
  \texttt{sbickel@northwell.edu}\\
  \And
  Ashesh D. Mehta \\
  The Feinstein Institutes for Medical Research\\
  Northwell Health\\
  Manhasset, NY 11030 \\
  \texttt{amehta@northwell.edu}\\
  \And
  Nima Mesgarani\thanks{Corresponding author} \\
  Department of Electrical Engineering\\
  Columbia University\\
  New York, NY 10027 \\
  \texttt{nima@ee.columbia.edu}
}

\begin{document}

\maketitle

\begin{abstract}
Recent advancements in artificial intelligence have sparked interest in the parallels between large language models (LLMs) and human neural processing, particularly in language comprehension. While prior research has established similarities in the representation of LLMs and the brain, the underlying computational principles that cause this convergence, especially in the context of evolving LLMs, remain elusive. Here, we examined a diverse selection of high-performance LLMs with similar parameter sizes to investigate the factors contributing to their alignment with the brain's language processing mechanisms. We find that as LLMs achieve higher performance on benchmark tasks, they not only become more brain-like as measured by higher performance when predicting neural responses from LLM embeddings, but also their hierarchical feature extraction pathways map more closely onto the brain’s while using fewer layers to do the same encoding. We also compare the feature extraction pathways of the LLMs to each other and identify new ways in which high-performing models have converged toward similar hierarchical processing mechanisms. Finally, we show the importance of contextual information in improving model performance and brain similarity. Our findings reveal the converging aspects of language processing in the brain and large language models and offer new directions for developing LLMs that align more closely with human cognitive processing.

\end{abstract}

\setcounter{figure}{0}

\section{Introduction}
The intersection of artificial intelligence and neuroscience has emerged as a frontier of great interest, particularly in understanding how large language models (LLMs) and the human brain process language. Prior research has laid foundational work in this area, uncovering intriguing parallels in feature extraction and representational similarities between LLMs and neural responses during language processing. Studies \cite{toneva2019interpreting, abnar2019blackbox, schrimpf2021neural, caucheteux2020language, hosseini2022artificial, anderson2021deep, caucheteux2021disentangling, caucheteux2022brains, sun2020neural} have demonstrated that the representations learned by LLMs can be linearly mapped to neural responses, suggesting that both LLMs and the brain utilize comparable features in language processing. However, these findings offer limited insight into the fundamental characteristics of LLMs that enable this brain-like processing.

Further investigations have delved into different aspects of LLMs to elucidate their resemblance to brain processes. Some studies \cite{goldstein2022shared, caucheteux2023evidence} support the predictive coding hypothesis in human language processing by finding stronger similarities with autoregressive LLMs. Others \cite{caucheteux2022brains, hosseini2022artificial, antonello2023scaling, antonello2023predictive} have explored various factors, such as the LLM language modeling performance, model size and capacity, and the generalizability of linguistic representations as indicators of brain-like processing. These studies imply that the quality of an LLM significantly contributes to its brain-like representations, yet the underlying reason for this similarity remain an open question. Is it merely a matter of scaling up the models \cite{antonello2023scaling}, or do these models share fundamental computational principles that increasingly align well with the spoken language processing pathway in the human brain? This question is significant as it may suggest a potential shift in the paradigm of model optimization. Although both brains \cite{hickok2007cortical, hasson2008hierarchy, lerner2011topographic, ding2017characterizing} and LLMs \cite{ethayarajh2019contextual, tenney2019bert} process speech and language in hierarchical pathways, most studies have analyzed the similarity of their representations without detailed comparisons of the hierarchical processes through which they are created. Thus, it is still unclear whether brains and models arrive at similar representations through the same or different pathways.

We aim to answer these questions by examining the interplay between LLM performance, neural predictability, anatomical alignment, and contextual encoding, potentially paving the way toward models that perform with high accuracy and process language in a manner similar to the human brain. We examined 12 open-source, pre-trained LLMs, all uniform in size but diverse in their linguistic capabilities, particularly in language-understanding tasks such as reading comprehension. We recorded neural responses with intracranial electroencephalography (iEEG) in the auditory cortex and speech processing regions of neurosurgical patients as they listened to speech. We then predicted these neural responses from the embeddings extracted from each layer of the LLMs as they processed the same linguistic input. This approach allowed us to pinpoint which layers and aspects of the LLMs were most predictive of brain activity and explore how variations in model performance align with differences in neural prediction and their anatomical and functional correspondence. Our findings offer a fresh perspective on the evolving landscape of LLMs, providing insights that reveal more intricate parallels in language comprehension between artificial and biological systems, uncover new potential reasons for LLM performance differences, and point to a convergence in LLMs towards a more optimal, brain-like language processing system.

\section{Results}
\subsection{Brain Similarity of Large Language Models}

\begin{figure}[!t]
  \centering
  {\includegraphics[width=0.95\linewidth]{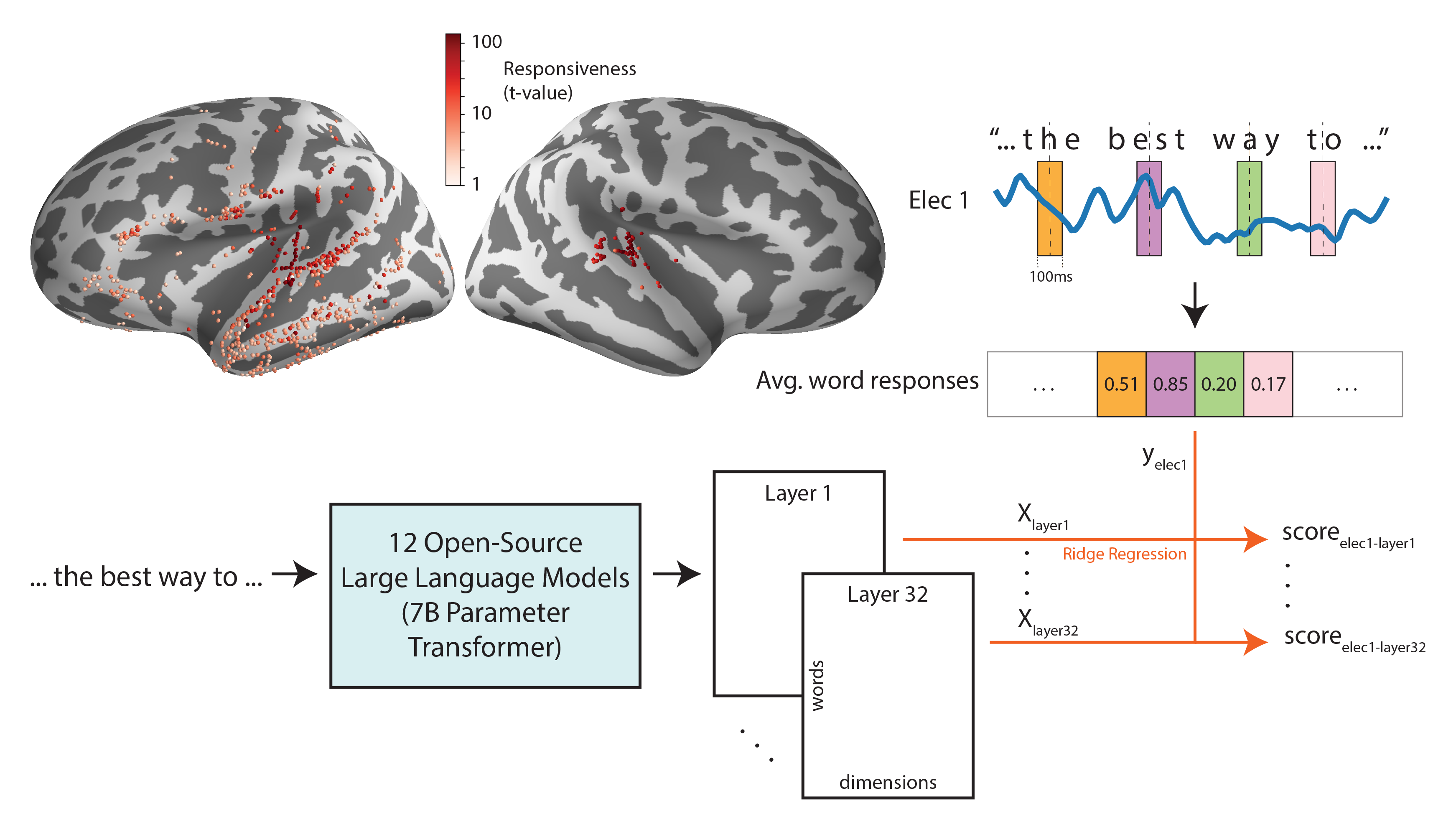}}
  \caption{Mapping LLM embeddings to the brain. Speech responsive electrodes are shown on an inflated brain (shaded by their responsiveness t-value from a paired t-test between speech and silence). As subjects listened to speech, the average neural response in a $100$ms window around a word center was used as a given electrode's word response. The same text was fed to an LLM and the embeddings from all $32$ layers were extracted. Ridge regression was used to predict the word responses from the LLM representations, producing a brain correlation score for each electrode-layer pair.}      
  \label{fig:1}
\end{figure}

We studied $12$ recent, popular, open-source LLMs, all with approximately $7$ billion parameters. We evaluated each model on a suite of benchmark tasks to assess its language modeling performance, splitting these tasks into categories relevant to English language comprehension, specifically reading comprehension and commonsense reasoning as in \cite{touvron2023llama2} (see Methods for details). Overall LLM performance was estimated as the average score over these two categories.

\begin{figure}[!t]
  \centering
  {\includegraphics[width=0.95\linewidth]{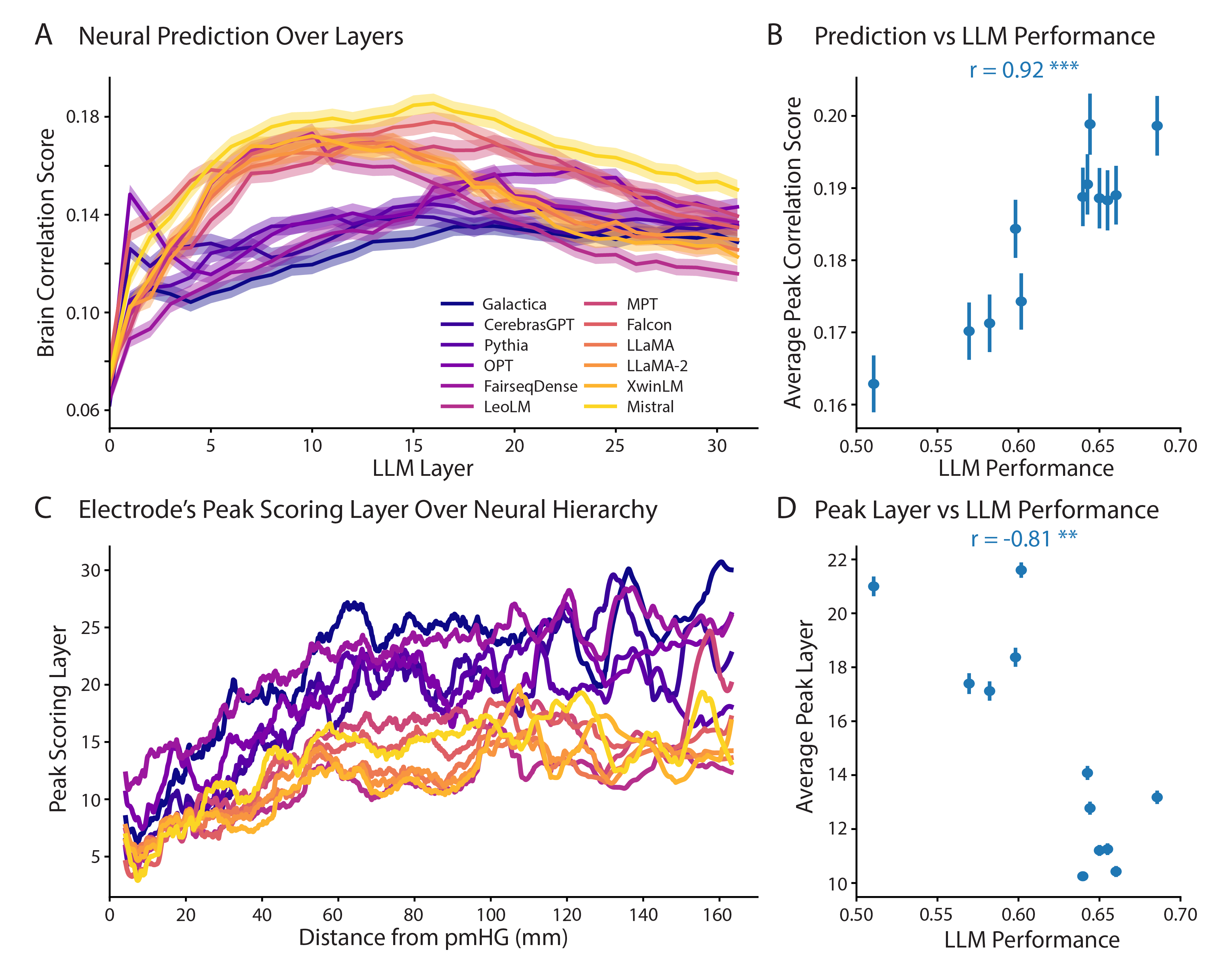}}
  \caption{Peak brain correlations and layers relate to LLM performance. A) Average brain correlation over all electrodes for each LLM. LLMs are colored in order of their separately-measured benchmark performance, with blue/purple models performing the worst and yellow models performing the best. Shaded regions indicate standard error of the mean over electrodes. B) The peak correlation over all layers of a given model was computed for each electrode, then averaged over all electrodes. Bars indicate standard error of the mean over electrodes. Average peak correlation score is significantly related to LLM performance (Pearson $r=0.92, p=2.24\times10^{-5}$). Stars indicate statistical significance level thresholds of $p<0.05$, $p<0.01$, and $p<0.001$ with *, **, and ***, respectively. C) The peak scoring layer of each model was computed for each electrode. Then electrodes were sorted by distance from pmHG and a sliding window average (centered, $n=50$) was taken across the electrodes of each model to compute the smoothed, local estimate of the most brain-like LLM layer. The peak scoring layer generally increases with distance from pmHG, and the better models (yellow) peak at lower layers compared to the worse models (blue/purple). D) The average peak layer for a given model over all electrodes is shown with bars indicating standard error of the mean. Average peak layer is significantly negatively related to LLM performance (Pearson $r=-0.81, p=0.0013$).}
  \label{fig:2}
\end{figure}

Neural responses were recorded with invasive electrodes (intracranial EEG) from eight neurosurgical patients, with electrode placement determined by clinical need (Supplementary Fig. 1). The subjects listened to between $20$ and $30$ minutes of speech from various talkers, including stories voiced by voice actors and dialogues between characters. The text for each audio was fed into each LLM, and we extracted the causal embeddings of each word at every layer. We reduced these embeddings to $500$ components with PCA, ensuring consistent dimensionality across models since the models were only approximately the same size to begin with. We used ridge regression to estimate the similarity of a model’s features to the brain (Fig. \ref{fig:1}) \cite{goldstein2022shared, schrimpf2021neural, caucheteux2022brains}. We analyzed $707$ electrodes which were responsive to speech, as determined by a t-test between responses to words and silence (FDR corrected, $p<0.05$ \cite{holm1979simple}). For each responsive electrode, we extracted the average high-gamma band envelope response in a $100$ms window around the center of every word. Then, we fit cross-validated ridge regression models to predict these neural responses from the word embeddings and used the average prediction correlation on the withheld folds as the brain similarity with that electrode. Neither the number of principal components of the embeddings nor the window size used to compute the neural response to words significantly impacted the results (Supplementary Fig. 2).

Electrode-averaged brain similarity over each model's layers is shown in Fig. \ref{fig:2}A. With these latest LLMs, we confirm previous findings showing that neural responses can be predicted from model representations, and we find that brain similarity generally increases over layers and peaks in middle or later layers \cite{schrimpf2021neural, caucheteux2022brains}. Higher-performing LLMs also achieve higher peak brain scores (Pearson $r=0.92, p=2.24\times10^{-5}$) (Fig. \ref{fig:2}B), indicating that they extract more brain-like features from language.

Similar to the layers of a model, the auditory and language processing pathway demonstrates hierarchical organization \cite{hickok2007cortical, sharpee2011hierarchical, hasson2008hierarchy, lerner2011topographic}. The primary auditory cortex, the first point of auditory processing in the cortex, is centered around posteromedial Heschl’s gyrus (pmHG, or TE1.1) \cite{morosan2001human}. Since this is a common reference point in auditory cortical processing, we quantify the depth of each electrode in the brain’s spoken language processing pathway using its distance from this landmark \cite{baumann2013unified, norman2018neural, mischler2023deep}. Prior studies have found that deeper layers of LLMs correspond better to deeper language processing regions of the brain \cite{caucheteux2022brains, caucheteux2023evidence, kumar2022reconstructing}. We confirm this result (Fig. \ref{fig:2}C), but interestingly, we also find that better-performing LLMs peak in brain similarity at earlier layers compared to worse models (Pearson $r=-0.81, p=0.0013$) (Fig. \ref{fig:2}D). This uncovers a new dimension in the evolution of LLMs: the progression of feature extraction over layers aligns differently with the brain for higher-performing versus lower-performing models.

\subsection{Alignment of Language Processing Hierarchies Between Models and the Brain}

\begin{figure}[!t]
  \centering
  {\includegraphics[width=0.95\linewidth]{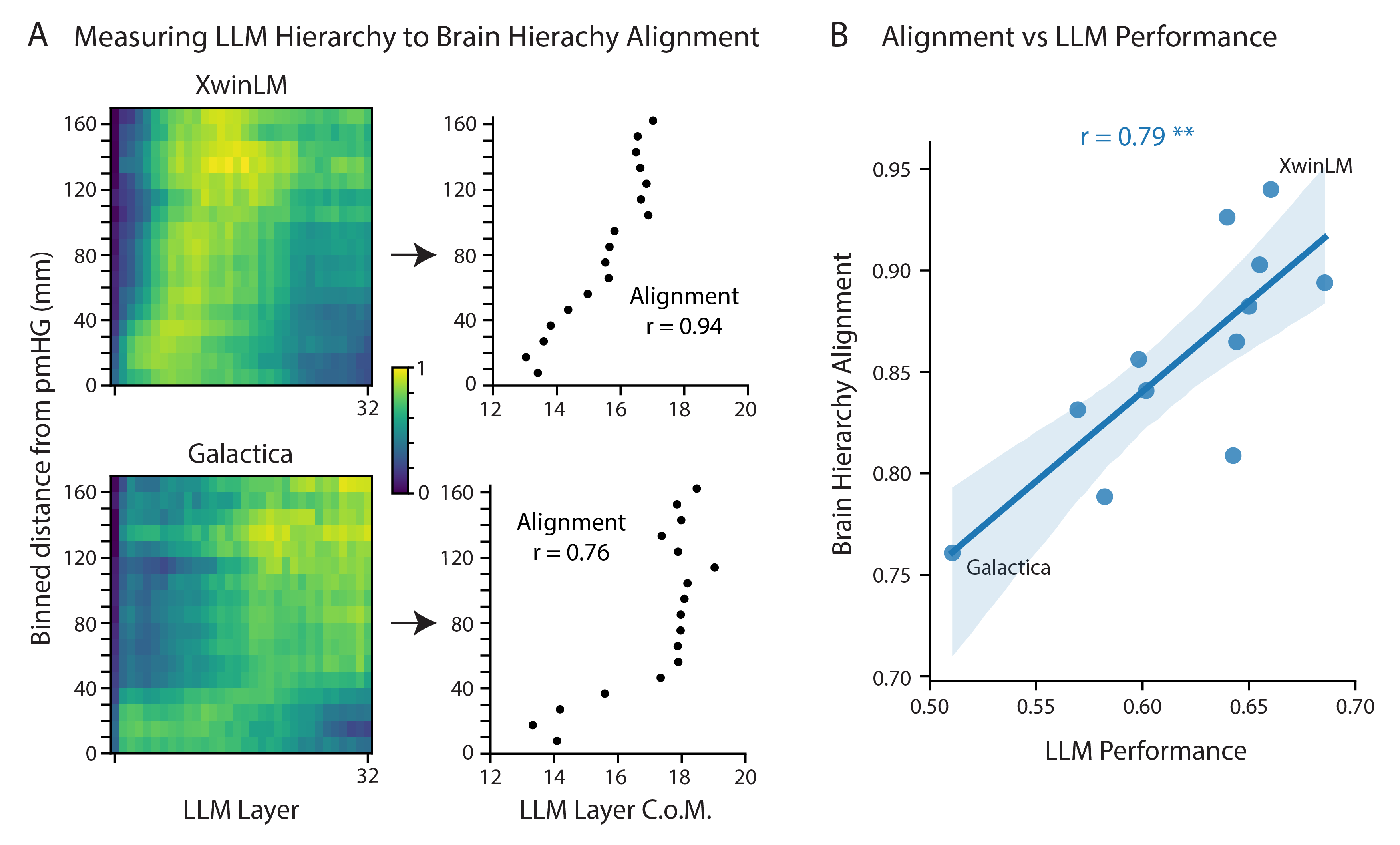}}
  \caption{Better LLMs display more brain-like hierarchical processing. A) Examples of computing the brain hierarchy alignment are shown for two models: XwinLM (the model with the highest alignment score) and Galactica (the model with the lowest alignment score). Electrodes were first binned into a hierarchy by distance from pmHG. Within a bin, the correlations over all 32 layers were normalized between 0 and 1 and then averaged over electrodes in the bin, producing one row for each bin in the matrix on the left. The center of mass (C.o.M.) of the distribution of brain similarity scores over LLM layers for each bin was computed and plotted in the scatter plot to the right. The brain hierarchy alignment score was then computed as the Pearson correlation between LLM layer C.o.M. and distance from pmHG. B) A scatter plot of brain hierarchy alignment scores and LLM performance shows a significant positive correlation (Pearson $r=0.79, p=0.0021$, ** indicates $p<0.01$). Line and shaded region shows linear regression fit and bootstrapped $(n=1000)$ $95\%$ confidence interval.}
  \label{fig:3}
\end{figure}

Given that the layer-wise brain similarity appears different between good and bad models, we hypothesized that better models were not only learning more brain-like features, but that the progression of feature extraction within these models was different. Taking inspiration from an investigation of hierarchical correspondence between stages of visual cortex processing and image classification networks \cite{nonaka2021brain}, we sought to compute the alignment between hierarchical feature extraction pathways in brains and models. Although the brain's exact hierarchical processing stages, analogous to layers of a model, are not perfectly known, we again used the distance from pmHG to quantify the stages of hierarchical processing. We grouped electrodes into bins at 10mm intervals. Then, for each electrode, we normalized the brain similarity scores over layers. Finally, we averaged these layer-wise scores over the electrodes in a bin, producing a set of layer scores which are shown as a single row of the alignment matrix in Fig. \ref{fig:3}A. We used the center of mass of this average brain similarity score over layers within each electrode bin to quantify the LLM layer most similar to a given stage of the brain's hierarchy. Then, we compared the progression of these most-similar LLM layers to the bin distances along the hierarchy, visually finding that some models achieve a more linear increase in LLM layers over bins. We summarize the alignment between the language processing hierarchies of each LLM and the brain using the Pearson correlation between the layer center of mass in each bin and the hierarchical stage of each bin (i.e. the distance of each bin from pmHG) \cite{nonaka2021brain}. We illustrate this alignment computation for XwinLM and Galactica, two models which achieve the highest and lowest hierarchy alignment scores, respectively (Fig. \ref{fig:3}A). These models also display a stark difference in benchmark performance, with Galactica being the lowest performing LLM. The alignment scores reveal that the better model (XwinLM) exhibits a feature extraction progression more consistent with the brain from early to late-stage processing compared to the bad model.  This brain alignment is also highly correlated with LLM performance on the benchmark evaluation tasks (Pearson $r=0.79, p=0.0021$) (Fig. \ref{fig:3}B). We find the same result when using electrode latency to measure the stages of the brain's hierarchical processing, rather than distance from pmHG (Pearson $r=0.89, p=0.0001$) (Supplementary Fig. 3), which demonstrates that this finding holds for other estimates of the stages of the cortical hierarchy. Additionally, to ensure this effect was not the result of a single subject overpowering the distribution, we separated the even- and odd-numbered subjects and performed the analysis again, finding that brain hierarchy alignment was significantly correlated with LLM performance for each group (Pearson correlation, even subjects: $r=0.79, p=0.0022$, odd subjects: $r=0.81, p=0.0013$) (Supplementary Fig. 4). Overall, these findings demonstrate that better-performing LLMs extract features using a hierarchy that more linearly aligns with the brain's hierarchical language processing pathway.

\begin{figure}[!t]
  \centering
  {\includegraphics[width=0.95\linewidth]{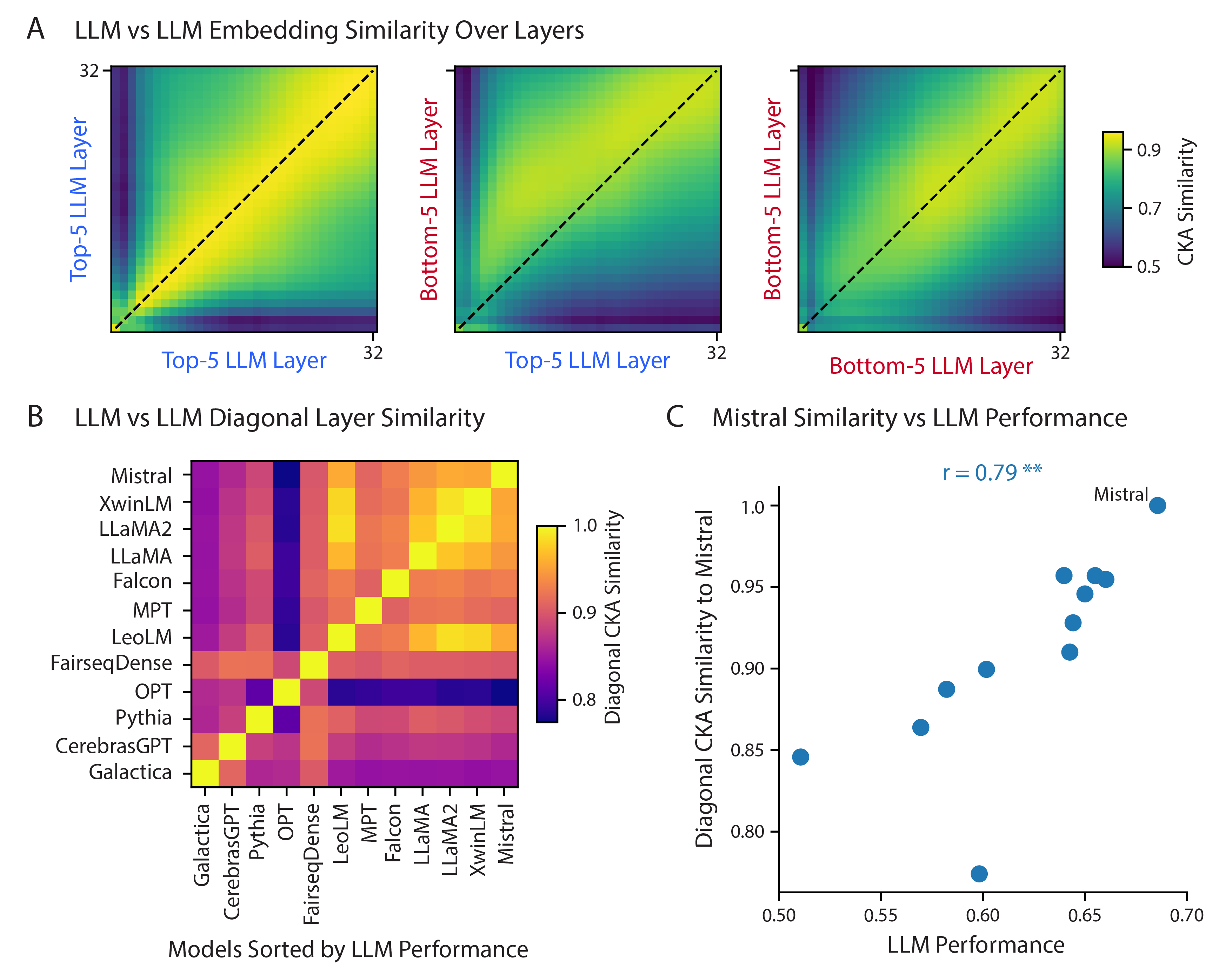}}
  \caption{Comparing feature extraction hierarchies between LLMs. A) Layer-by-layer similarity matrices were computed using CKA for every pair of LLMs. LLMs were labeled as either top-5, bottom-5, or excluded, depending on their sorted performance on our LLM benchmark evaluation. Then, similarity matrices between all pairs of top-5 LLMs were averaged and displayed as the ``top-5 versus top-5'' average similarity matrix in the top left. The same was done to create similarity matrices between the average ``top-5 versus bottom-5'' as well as the average ``bottom-5 versus bottom-5'' LLMs. Visually, the ``top-5 versus top-5'' similarity matrix is highly diagonal, while ``bottom-5 versus bottom-5'' is less similar in early layers. The ``top-5 versus bottom-5'' similarity matrix shows an offset diagonal, indicating a delay in feature extraction for the bottom-5 models compared to the top-5. B) Model-by-model diagonal similarity was computed as the average along the diagonal of their similarity matrix. Models are arranged in sorted order of LLM benchmark performance from worst to best. This visually confirms that the best models are fairly similar to each other in layer-wise feature extraction, while worse models are less similar to each other and less similar to the best models. C) The diagonal similarity of each model with Mistral, the best performing LLM, is plotted against the LLM performance, showing a strong positive relationship (Pearson $r=0.79, p=0.0022$, ** indicates $p<0.01$).}
  \label{fig:4}
\end{figure}

To perform model-to-model comparisons, we used centered kernel alignment (CKA) \cite{kornblith2019similarity}, a method analogous to canonical correlation analysis (CCA) with a nonlinear kernel, which is able to capture similarity between high-dimensional representations like neural network embeddings. We computed the CKA similarity between all pairs of layers for all pairs of models. Thus, each pair of models creates a layer-by-layer similarity matrix describing their embedding similarity. High similarity along the diagonal indicates that the two models extract similar features at the same layers. Higher similarity offset from the diagonal indicates that one model exhibits a delay in extracting similar features. When grouping these similarity matrices by the top-5 and bottom-5 models based on LLM benchmark performance and averaging within a group, an interesting pattern emerges (Fig. \ref{fig:4}A). We find that the top models exhibit a high degree of similarity to each other along the diagonal. On the other hand, the worst models are much less similar to each other in their early layers, and even in their later layers they are less consistent than the top-5-to-top-5 model pairs. Finally, comparing top-5 models to bottom-5 models reveals a striking offset in maximum similarity from the diagonal. This suggests that bad models require more layers to reach a similar level of feature extraction as good models. We summarize the layer-wise feature extraction similarity between each pair of models using the average CKA similarity along the diagonal in their CKA similarity matrix (Fig. \ref{fig:4}B). The plot demonstrates that the top-5 models are indeed more similar to each other, with a sub-block of high similarity emerging among the top few models. Since Mistral is the best performing LLM, we look at the diagonal similarity to Mistral of each model and find that a more Mistral-like feature extraction progression correlates strongly with LLM performance (Pearson $r = 0.79, p=0.0022$) (Fig. \ref{fig:4}C). These results reveal new distinctions between the embeddings of LLMs and suggest that inefficient feature extraction or poor early-layer learning in bad models may contribute to their worse performance and lower brain similarity.

\subsection{Contextual Content Supports Brain Hierarchy Alignment}

\begin{figure}[!t]
  \centering
  {\includegraphics[width=0.95\linewidth]{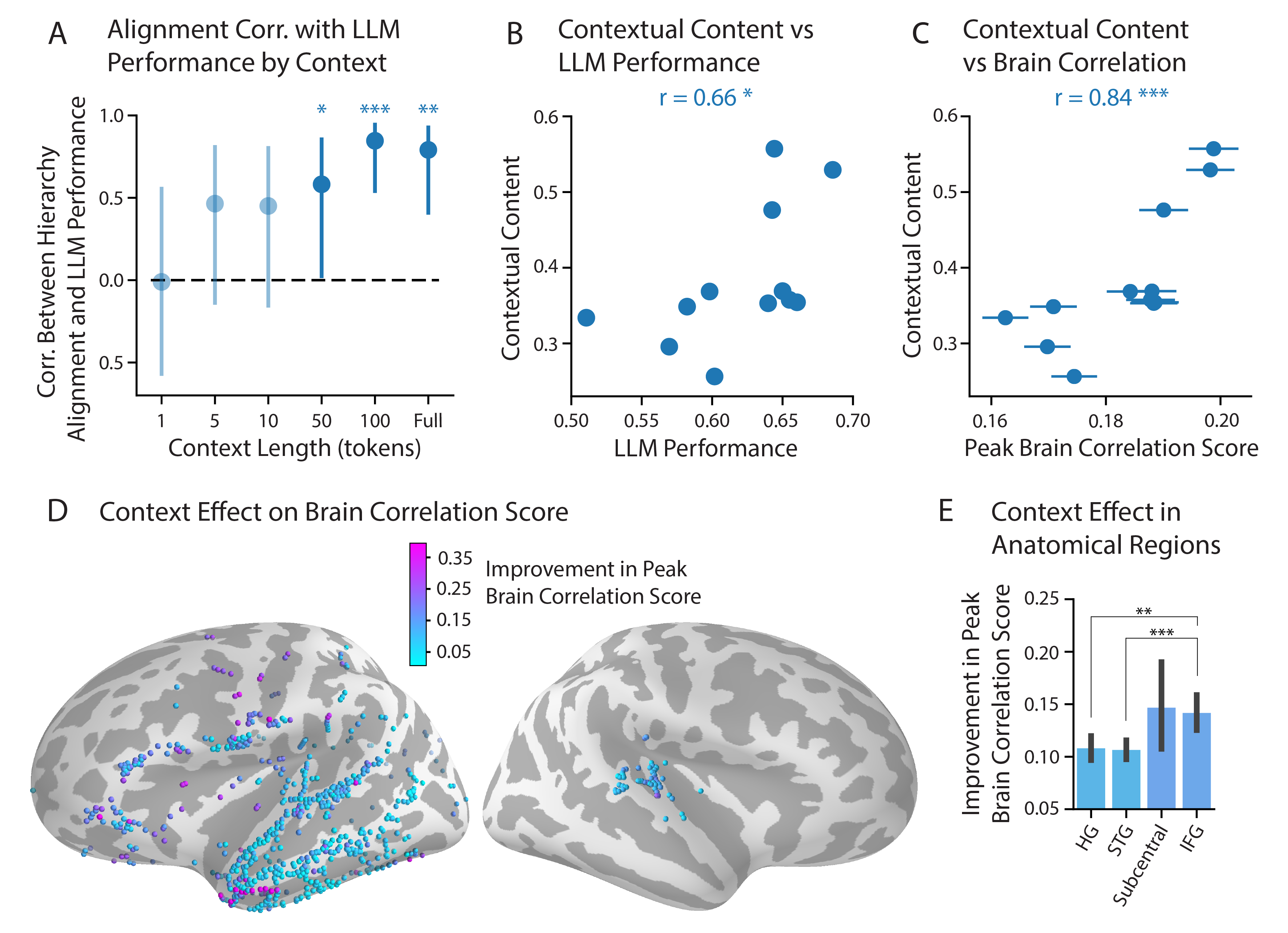}}
  \caption{Effect of contextual information. A) Using embeddings from the LLMs when given a certain limited number of the previous tokens as a context window, we performed the analysis of brain hierarchy alignment again. The correlation between LLM performance and brain hierarchy alignment is illustrated by each dot with 95\% confidence interval bars, showing only a significant correlation for long contextual windows. Stars illustrate the significance level of the correlation, with $*$, $**$, and $***$ indicating chance levels below $0.05$, $0.01$, and $0.001$, respectively. B) The contextual content of a model's representations is plotted against its benchmark performance, showing a positive correlation between the two (Spearman $r=0.66, p=0.020$). C) Contextual content of each model is plotted against its average peak brain similarity over electrodes, showing a strong correlation (Spearman $r=0.84, p=0.0006$). Horizontal lines show standard error of the mean over electrodes for brain similarity. D) Electrodes plotted on the FreeSurfer average inflated brain, colored by the effect of contextual information on peak brain similarity score. E) Bar plot of the average context effect on peak brain similarity score for electrodes within four main anatomical regions along the linguistic hierarchy. Each bar is colored by its value according to the same colormap as used on the brain plot, and error bars show standard error of the mean. Stars indicate significant differences between a pair of regions (Wilcoxon rank-sum test).}
  \label{fig:5}
\end{figure}

Since the contextual nature of LLM features is critical for their brain similarity compared with non-contextual representations \cite{goldstein2022shared, schrimpf2021neural, caucheteux2022deep}, we hypothesized that the amount of contextual information used by a model may also play a key role in determining the alignment between hierarchical feature extraction pathways of LLMs and the brain. We extracted limited-context embeddings from the LLMs by restricting their causal attention mechanism to a certain window of the previous text. Transformer architecture LLMs use tokenizers to separate text into discrete units, so we supplied the models with only the most recent $N$ tokens, sweeping $N$ over a range of values from $1$ to $100$. A single token input gives the model no context at all. For reference, Mistral's tokenizer averages $1.15$ tokens per word in our stimulus corpus. We then repeated our analysis of model-brain hierarchical alignment (as previously shown in Fig \ref{fig:3}B) by computing the correlation between brain hierarchy alignment and LLM performance at each limited context window length. While this correlation is positive for all but the $1$-token case, it is only significant for long contextual window lengths of $50$ tokens and above (Fig. \ref{fig:5}A). This suggests that the brain alignment of LLMs critically depends on the amount of contextual information the model is able to see, which then influences its hierarchical feature extraction mechanism.

Since the correlation between LLM performance and hierarchical alignment is strongly positive for long context lengths, we expected that better-performing models would be better at incorporating contextual information into their language representations. To test this, we quantified the amount of contextual information present in a model's embeddings by measuring how much its embeddings changed when contextual information was added to the input. We measured the CKA difference ($1 - \text{similarity}_{CKA}(\text{full-context}, \text{1-token})$) of the embeddings of each layer when given the full context compared to the first-layer embeddings when given only a $1$-token limited context window. We refer to the average of this CKA difference over all layers as the contextual content of the model's representations. We find that this contextual content is positively correlated with LLM performance (Spearman $r=0.66, p=0.020$) (Fig. \ref{fig:5}B). Additionally, it is very strongly correlated with brain similarity (Spearman $r=0.84, p=0.0006$) (Fig. \ref{fig:5}C). These findings indicate that contextual information plays a crucial role in natural language processing in both natural and artificial language models, and contextual feature extraction enables brain hierarchy alignment in LLMs.

We further investigated the impact of contextual information on neural similarity by computing how much each LLM's peak similarity score with a given electrode changed when the models were given the full context versus no context ($1$ token). 
We then averaged this difference over all LLMs for each electrode and plotted the electrodes on the brain, finding that being given the extra context more greatly improved similarity scores with electrodes in higher-level language processing areas (Fig. \ref{fig:5}D). Averaging electrodes within major anatomical regions further quantifies this result, as we find higher average context effects on brain correlation score within the higher-level linguistic-processing area of inferior frontal gyrus (IFG) \cite{costafreda2006systematic} compared to sensory regions like Heschl's gyrus (HG) and superior temporal gyrus (STG) (Wilcoxon rank-sum test, $p<0.05$) (Fig. \ref{fig:5}E). Interestingly, the articulatory region of subcentral gyrus, which has also been implicated in high-level linguistic processing \cite{arana2020sensory}, displays the highest average score improvement, but due to its high variance it does not meet statistical significance.
These results show that contextual information becomes more critical in determining brain similarity further along the spoken language processing hierarchy, which supports previous investigations of high-level linguistic feature encoding in more downstream regions \cite{sheng2019cortical, keshishian2023joint}. 
This finding strengthens the notion that both the brain and LLMs are extracting context along their hierarchies, and that LLMs need contextual information to achieve brain similarity in downstream processing regions. Taken together, our analyses reveal that high-performing LLMs not only extract representations of language that are similar to the brain, but they also use hierarchical feature extraction pathways which more strongly align with that of the brain due to contextual information processing abilities, a finding that uncovers new ways in which the best LLMs are continuously converging toward the brain.

\section{Discussion}

We explored LLMs and their alignment with neural responses during language processing, uncovering several key findings. Firstly, we observed a clear correlation between the language task performance of LLMs and their accuracy in predicting neural responses in the auditory cortex, with higher-performing models exhibiting greater functional alignment with the speech cortex. Secondly, we showed that the models with higher performance on benchmark tasks achieved peak predictive accuracy in earlier layers. In contrast, lower-performing models exhibited a delayed representation, necessitating deeper layers to approach similar levels of brain prediction accuracy. Finally, our study highlights the crucial role of contextual information in both LLMs and brain processing, where the contextual window's size significantly influenced the difference between better and worse models, with the availability of long-range contextual information driving the high-performing LLMs closer to the brain's hierarchical pathway. These findings uncover fundamental principles in language processing, highlighting the critical role of hierarchical structure and contextual dependencies in language which give rise to convergent processing strategies in both artificial and biological systems. 

\subsection{Hierarchical Processing and Inter-Model Comparisons}
We found that better-performing LLMs exhibit a more brain-like hierarchy of layers, offering new insights into their language processing. While previous studies have revealed similarities in the hierarchical stages found in the brain and deep neural networks for linguistic \cite{caucheteux2023evidence, caucheteux2022brains, kumar2022reconstructing}, acoustic \cite{giordano2023intermediate, tuckute2023many}, visual \cite{kriegeskorte2015deep, cichy2016comparison, sexton2022reassessing}, and imagined stimuli \cite{horikawa2017hierarchical}, a distinct approach in our study is the inter-model comparison within a consistent architectural framework. In related work analyzing deep neural networks for vision tasks, recent evidence \cite{nonaka2021brain} has shown that better performance can create a less brain-like progression of feature extraction in models when compared to the visual cortex, suggesting that the complex architectures of high-performing image processing networks have steered them away from neural alignment. By examining LLMs based on a single architecture, the stacked transformer decoder \cite{vaswani2017attention}, we uncover differences in their alignment with the brain's hierarchical stages during language comprehension. Transformer language models use contextual features to encode linguistic, syntactic, and positional structures \cite{o2021context, clark2019does}, and increasingly high-level and context-specific features arise throughout a model’s layers \cite{ethayarajh2019contextual, tenney2019bert}. This may be partly because later layers bind linguistic structures over longer contexts \cite{skrill2023large}. The crucial observation that such models display brain-like hierarchies resonates with neurobiological findings of hierarchical organization in the auditory and language-related cortex \cite{hickok2007cortical, sharpee2011hierarchical, sheng2019cortical, ding2017characterizing, hasson2008hierarchy, lerner2011topographic, norman2022multiscale, de2017hierarchical}. The convergence of the two systems highlights language's inherent hierarchical structure as we increasingly form larger units of representation, from articulatory features to phonemes, syllables, words, sentences, and phrases \cite{keshishian2023joint, di2021neural, gong2023phonemic}. Our results demonstrate that as LLMs have achieved higher performance, they have done so using feature extraction pathways that more closely resemble the human brain.

\subsection{Feature Extraction Efficiency and Contextual Processing}
A significant finding of our study is the delayed feature extraction observed in less effective LLMs compared to their higher-performing counterparts. This delay, particularly evident in the early processing stages within transformer models, suggests a slower buildup of relevant linguistic and contextual information \cite{tenney2019bert}. The implications of this observation are multifaceted. Firstly, it challenges the conventional emphasis on the final layers of LLMs \cite{goldstein2022shared}, instead drawing attention to the critical role of initial layers in efficient language processing \cite{antonello2023predictive}. This shift in focus aligns with emerging neuroscience research that underscores the significance of early-stage processing in the human brain for complex cognitive tasks like language processing \cite{de2017hierarchical, keshishian2023joint, gong2023phonemic}. Secondly, this delayed representation in less effective models offers insights into potential inefficiencies in their training or design. Given the architectural similarity of models in our study, the variance in feature extraction efficiency among models may reflect differences in training strategies \cite{naveed2023comprehensive} and data quality \cite{raffel2020exploring, lee2021deduplicating, touvron2023llama2}, providing insights for future LLM model development. As LLMs have evolved in recent years, improvements in dataset size and cleanliness as well as architectural changes to increase context length have come along with their performance improvements, and our results show that these improvements have also given rise to greater brain similarity. Furthermore, the observation that higher-performing models utilize early layers more effectively and peak in their brain similarity in middle layers rather than later layers raises intriguing questions about the role of subsequent layers. It is possible that these later layers are engaged in next-level contextual integration and feature extraction, potentially analogous to higher-order stimulus integration to support cognitive functions in the human brain \cite{huth2016natural, murphy2023spatiotemporal}. Alternatively, this finding could point to a limitation in our current methodologies, such as limited iEEG coverage, the simplicity of the speech comprehension task, or the fact that LLMs are not explicitly trained to perform comprehension, but rather next-word prediction, which is slightly different from the speech listening comprehension task the subjects performed. Our iEEG recordings include broad coverage of speech processing regions, especially acoustic sensory regions like HG and STG, which, although critical for spoken language processing, represent a slightly different aspect of linguistic feature extraction than the token-level processing that transformer architecture LLMs begin with. Answering these questions is crucial for enriching our understanding of artificial language processing.

The influence of contextual information on brain similarity and LLM benchmark scores also points to specific avenues that may improve model performance on language tasks. Ensuring that models are able to extract long context windows, such as by using architectures that allow for long context windows \cite{xiong2023effective} and utilizing training data that is rich in long context information, could enhance LLM performance further beyond simply scaling up a model's parameter size. Transformer-based LLMs have been shown to suffer from unequal contextual information extraction when the prior context occurs at different distances from the target \cite{liu2023lost}, supporting the notion that improving the robustness of modern LLMs to varying context lengths may lead to performance improvements. Our investigation offers a unique lens through which to view the parallels and divergences between machine learning and human cognitive development.

\subsection{Convergence to Brain-Like Models for Human-Level Artificial General Intelligence}

The convergence of LLMs and human speech processing may suggest that certain fundamental principles underlying efficient language processing might be common to both artificial and biological systems. The human brain's language capabilities have developed as an adaptive response to complex communication needs, optimizing for efficiency and versatility \cite{pinker1990natural}. Our findings suggest that LLM architectures and processing strategies are gravitating towards these same principles, mimicking the brain’s evolutionary adaptations for language. LLMs are trained without consideration for brain similarity, yet they have become increasingly brain-like in their feature extraction and hierarchical processing. Brain-like processing may represent an optimal solution to language modeling found by evolution \cite{deacon1997symbolic}, although subject to biological constraints, and our results suggest that modern LLM training focused on performance optimization may have placed these models on a similar path. In our study, Mistral, the top-performing model, stands as a prime example of this convergence, where the degree of similarity of a model’s embeddings to those of Mistral is highly correlated with performance and brain similarity. This evolution towards an optimal brain-like model offers an intriguing suggestion regarding artificial general intelligence (AGI). While not clearly defined, AGI can be quantified as human-level performance on a broad set of benchmarks \cite{goertzel2014artificial}. Our findings suggest that developing models mimicking human neural processing strategies \cite{zhao2023brain}, rather than solely focusing on augmenting computational power or diversifying learning algorithms \cite{zhao2023survey}, could accelerate the development of models that behave on par with human performance. Hence, brain similarity could be a useful evaluation and optimization metric for future model development.

Our research marks a significant stride in understanding the parallels between large language models and human brain processes in language comprehension, by revealing the intricate relationship between internal model representation, model performance, and neural predictive accuracy. Our findings enhance the understanding of LLMs and offer new insights into the cognitive mechanisms underlying human language processing.

\section{Methods}
\subsection{Human Intracranial Recordings}

Eight subjects undergoing clinical evaluation for drug-resistant epilepsy participated in the study. Electrodes were implanted intracranially (iEEG) with the clinical goal of identifying epileptogenic foci for surgical removal. Any electrodes showing signs of epileptiform discharges, as identified by an epileptologist, were not analyzed in this study. Prior to electrode implantation, all subjects provided written informed consent for research participation. The research protocol was approved by the institutional review board at North Shore University Hospital.

Subjects listened to naturalistic recordings of voice actors reading passages from stories and conversations. To ensure the subjects were paying attention to the stimuli, one of the voices in the recording occasionally directed a question at the listener directly, or the stories were paused and the subject was asked a question, to check their understanding. The subjects were able to effectively answer each question. These pauses separated the stimulus into separate passages.

The envelope of the high-gamma band (70-150 Hz) of the raw neural recordings was computed using the Hilbert transform \cite{edwards2009comparison} and downsampled to 100 Hz. This signal was used as the neural response due to its correlation with neuronal firing rates \cite{ray2011different, steinschneider2008spectrotemporal} and its common use in auditory neuroscience research \cite{mesgarani2014phonetic, bouchard2013functional}. We restricted our analysis to speech-responsive electrodes, which we estimated using a t-test between each electrode's response to the first second of the stimulus compared to last second of silence preceding it (FDR corrected, $p<0.05$ \cite{holm1979simple}), which left $707$ electrodes for analysis. We extracted average word responses from each electrode by taking the average high-gamma signal value in a 100ms window around the midpoint of each word.

\subsection{Large Language Models}

We analyzed $12$ LLMs of approximately $7$ billion parameters downloaded from Hugging Face and implemented with its \texttt{Transformers} library \cite{wolf2019huggingface}, including the most recent and most popular open-source LLMs. We selected these models by searching the Hugging Face Hub for 7 billion parameter models, then using as many of the trending or most-downloaded models that we were able to run without issue. 

We computed two similar evaluation metrics to those used by LLaMA 2 \cite{touvron2023llama2}: Reading Comprehension and Commonsense Reasoning. As measures of English language understanding, these are both highly related to the listening comprehension task which was performed by the human subjects in the study. As in \cite{touvron2023llama2}, these metrics were created by averaging the model's performance on a certain set of related tasks. All individual benchmarks were computed for each model using the Language Model Evaluation Harness \cite{evalharness} on Github.

\begin{itemize}
  \item Reading Comprehension - This metric was the average 0-shot performance of a model on SQuAD 2.0 \cite{rajpurkar2018know} and BoolQ \cite{clark2019boolq}.
  \item Commonsense Reasoning - This metric consists of the average 0-shot performance on OpenBookQA \cite{mihaylov2018can}, PIQA \cite{bisk2020piqa}, HellaSwag \cite{zellers2019hellaswag}, and WinoGrande \cite{sakaguchi2021winogrande}.
\end{itemize}

Overall LLM Performance was computed as the average Reading Comprehension and Commonsense Reasoning scores.

The models used, and their benchmark performance and overall LLM performance scores, are shown in Table \ref{table:models}.


\begin{table}[!h]
\begin{center}
\begin{tabular}{||c | c | c | c||} 
 \hline
 Models Used &  \multicolumn{1}{|p{2.5cm}|}{\centering Reading \\ Comprehension} & \multicolumn{1}{|p{2.5cm}|}{\centering Commonsense \\ Reasoning}  & LLM Performance \\ [0.5ex] 
 \hline\hline

 Galactica-6.7B \cite{taylor2022galactica} & 0.486 & 0.535 & 0.511 \\
 \hline
 CerebrasGPT-6.7B \cite{dey2023cerebras} & 0.565 & 0.575 & 0.570 \\
 \hline
 Pythia-6.9B \cite{biderman2023pythia} & 0.568 & 0.597 & 0.582 \\
 \hline
 OPT-6.7B \cite{zhang2022opt} & 0.581 & 0.616 & 0.598 \\
 \hline
 FairseqDense-6.7B \cite{artetxe2021efficient} & 0.575 & 0.628 & 0.602 \\
 \hline
 LeoLM-7B \cite{leo2023leohessianai} & 0.634 & 0.646 & 0.640 \\
 \hline
 MPT-7B \cite{mosaic2023introducing} & 0.620 & 0.665 & 0.643 \\
 \hline
 Falcon-7B \cite{almazrouei2023falcon} & 0.619 & 0.669 & 0.644 \\
 \hline
 LLaMA-7B \cite{touvron2023llama} & 0.626 & 0.674 & 0.650 \\
 \hline
 LLaMA2-7B \cite{touvron2023llama2} & 0.639 & 0.671 & 0.655 \\
 \hline
 XwinLM-7B \cite{xwin2023xwin} & 0.648 & 0.673 & 0.660 \\
 \hline
 Mistral-7B \cite{jiang2023mistral} & 0.669 & 0.703 & 0.686 \\
 \hline
\end{tabular}
\caption{All models used in the study, along with their computed benchmark performances.}
\label{table:models}
\end{center}
\end{table}


In order to extract LLM embeddings for each stimulus passage (approximately 30-60 seconds when spoken), we fed the text to the model and extracted the embeddings of each layer when given a causal attention mask. When limiting the contextual window of the model, the attention mask was truncated to only include the most recent $N$ tokens. For multi-token words, we used the embedding of the last token in the word. Thus, for each passage, we extracted a tensor of embeddings of shape $(L_{layers}, N_{words}, D_{dimensions})$ from each model.





\subsection{Ridge Regression Mapping from Embeddings to Neural Responses}

We performed PCA to reduce the dimensionality of each model's embeddings to 500 components. For a given model, PCA was performed for each layer separately. Then, we fit $10$-fold cross-validated ridge regression models to predict the average word responses from each layer's embeddings, sweeping over a range of regularization parameters for each training fold, using \texttt{scikit-learn}'s \texttt{RidgeCV} model \cite{pedregosa2011scikit}.

\subsection{Electrode Localization and Brain Plotting}

Each subject's electrode positions were mapped to the subject's brain using \texttt{iELVis} \cite{groppe2017ielvis} to perform co-registration between pre- and post-implant MRI scans. Then, the subject-specific electrode locations were mapped to the FreeSurfer average brain \cite{fischl2004automatically}. Euclidean distance from posteromedial HG (TE1.1) \cite{morosan2001human} was computed in this average brain, since TE1.1 is a landmark of primary auditory cortex \cite{baumann2013unified, norman2022multiscale, norman2018neural, mischler2023deep}. When visualizing electrodes on the average brain, all subdural electrodes were snapped to the nearest surface point.

\subsection{Comparing LLMs with Centered Kernel Alignment}

To estimate the similarity between high-dimensional embeddings of different models, we used CKA \cite{kornblith2019similarity}, a similarity metric which is related to CCA but has been shown to perform well in high-dimensional scenarios between neural network features. We used the RBF kernel to allow for nonlinear similarity measurement. For a given pair of models, we computed the CKA similarity between the embeddings of one layer of the first model with another layer of the second model. Iterating over all pairs of layers for those two models produced a single similarity matrix. These similarity matrices were then grouped by whether they described a comparison between two models in the top-5 of all LLMs for benchmark performance, one model in the top-5 and the other in the bottom-5, or two models in the bottom-5, and then averaged.

\subsection{Data and Code Availability}

Although the iEEG recordings used in this study cannot be made publicly available, they can be requested from the author [N.M.]. Code for preprocessing neural recordings, including extracting the high-gamma envelope and identifying responsive electrodes is available in the \texttt{naplib-python} package \cite{mischler2023naplib}.

\section*{Acknowledgement}
This work was funded by the National Institutes of Health, the National Institute on Deafness and Other Communication Disorders, and the National Science Foundation Graduate Research Fellowship Program. The funders had no role in study design, data collection and analysis, decision to publish or preparation of the manuscript.

\bibliography{main}
\bibliographystyle{unsrt}

\newpage

\section{Supplementary Figures}
\renewcommand{\figurename}{Supplementary Figure}
\setcounter{figure}{0}

\begin{figure}[ht]
  \centering
  {\includegraphics[width=0.95\linewidth]{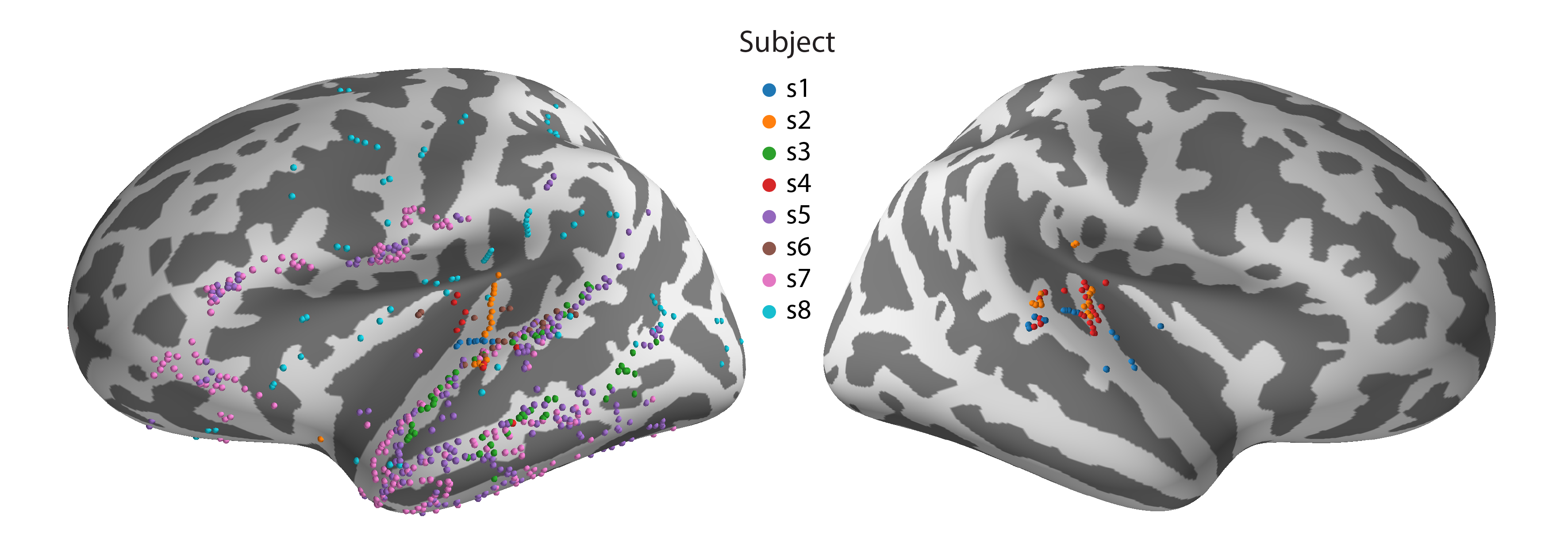}}
  \caption{Subject-wise electrode localization. Electrodes are plotted on the inflated Freesurfer average brain and are colored by their corresponding subject identity.}
  \label{fig:s1}
\end{figure}

\begin{figure}[ht]
  {\includegraphics[width=0.47\linewidth]{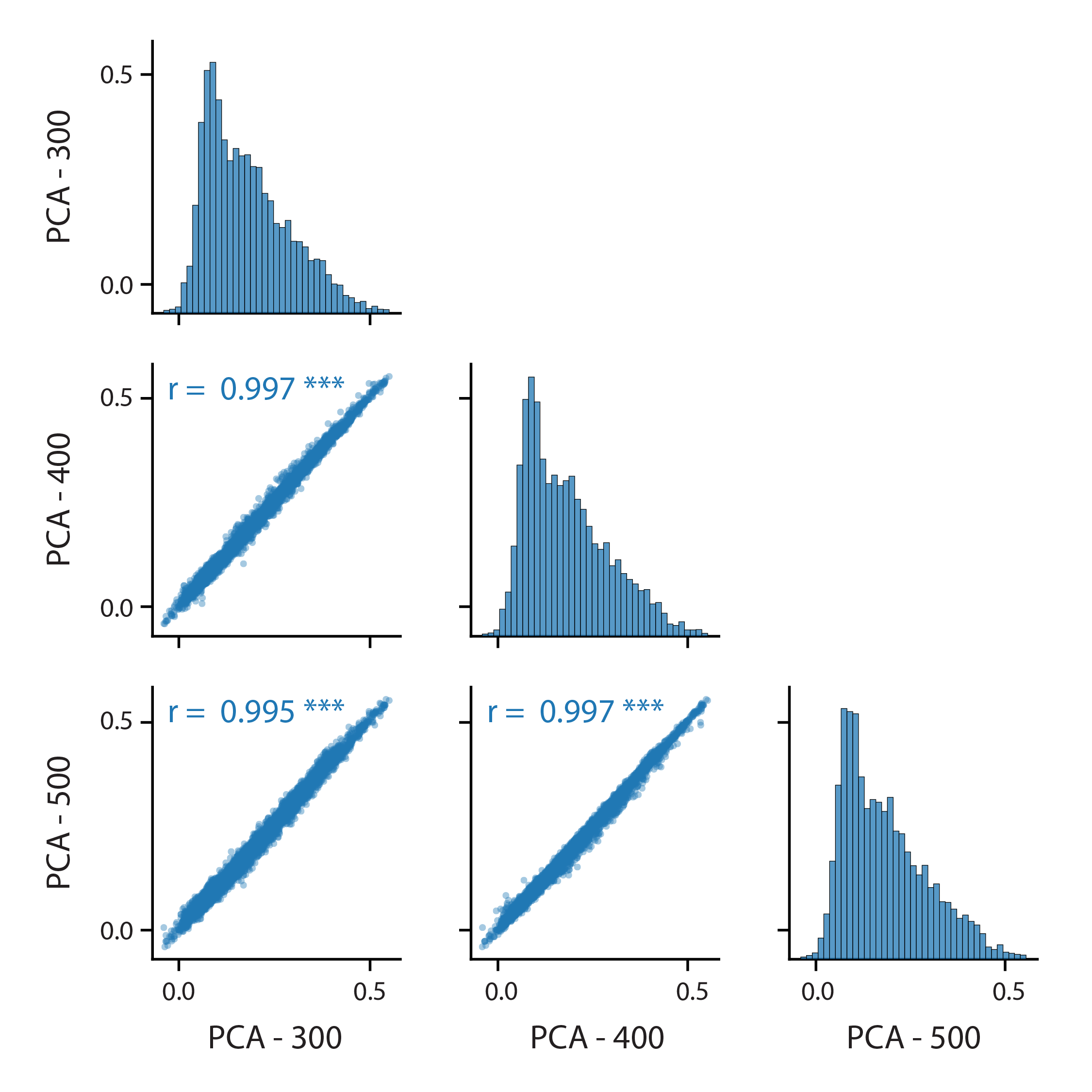}}
  {\includegraphics[width=0.47\linewidth]{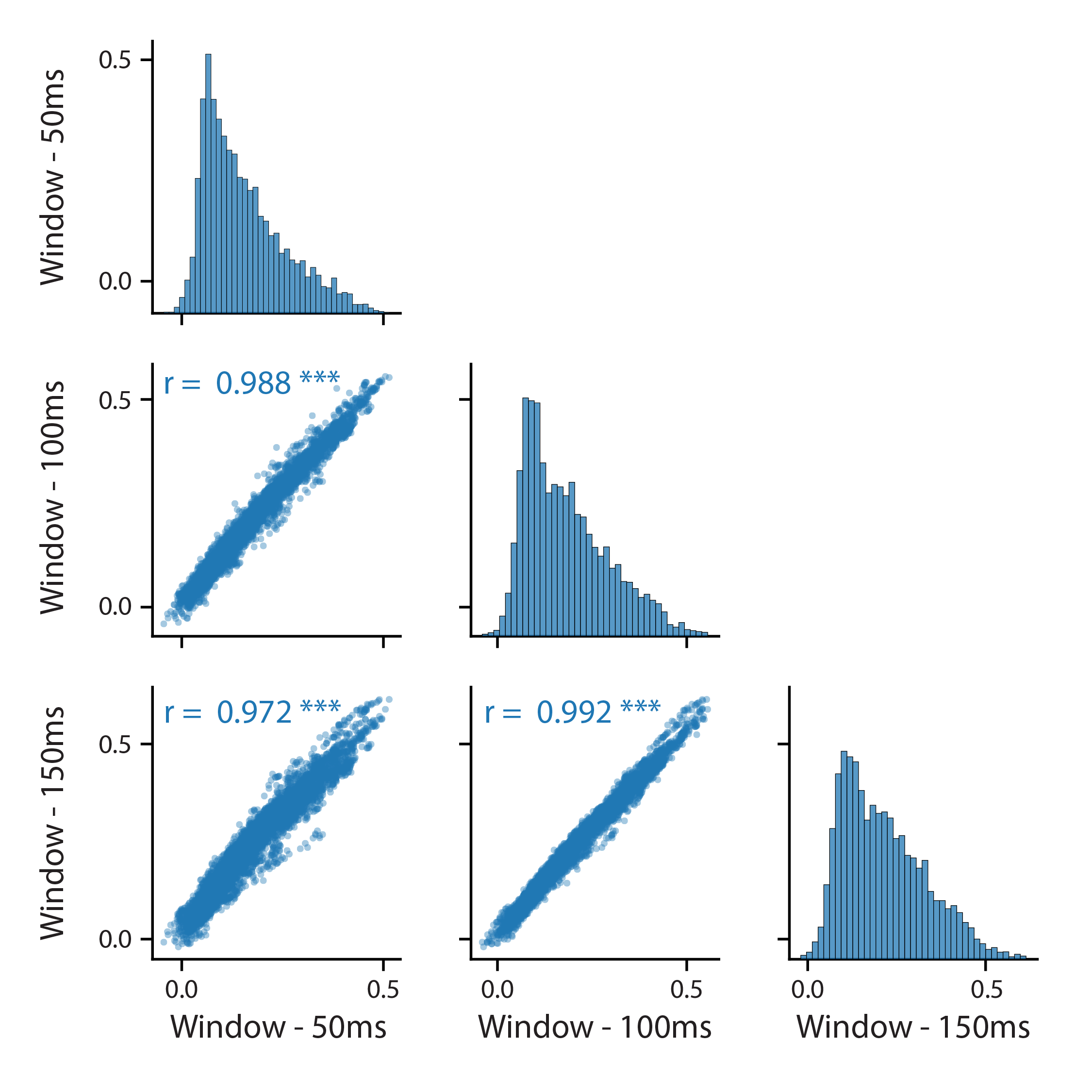}}
  \caption{Effect of regression hyperparameters on scores. The left plot shows the pairwise effects on the peak brain similarity scores when altering the number of principal components of the LLM embeddings used for computing scores with ridge regression, keeping a $100$ms window size constant. The right plot shows the pairwise effects of altering the width of the averaging window around word centers for estimating neural responses to words, keeping the PCA dimensionality of $500$ constant. Along each plot’s diagonal is the marginal distribution for that hyperparameter setting. The off-diagonal plots display scatter plots of all the peak-scores for all models together for one hyperparameter setting against another. Each dot represents the peak brain correlation score for one model-electrode pair. All pairs of settings produce scores which are highly correlated, as written in each subplot (Pearson correlation, *** indicates $p<0.001$).}
  \label{fig:s2}
\end{figure}

\begin{figure}[ht]
  \centering
  {\includegraphics[width=0.45\linewidth]{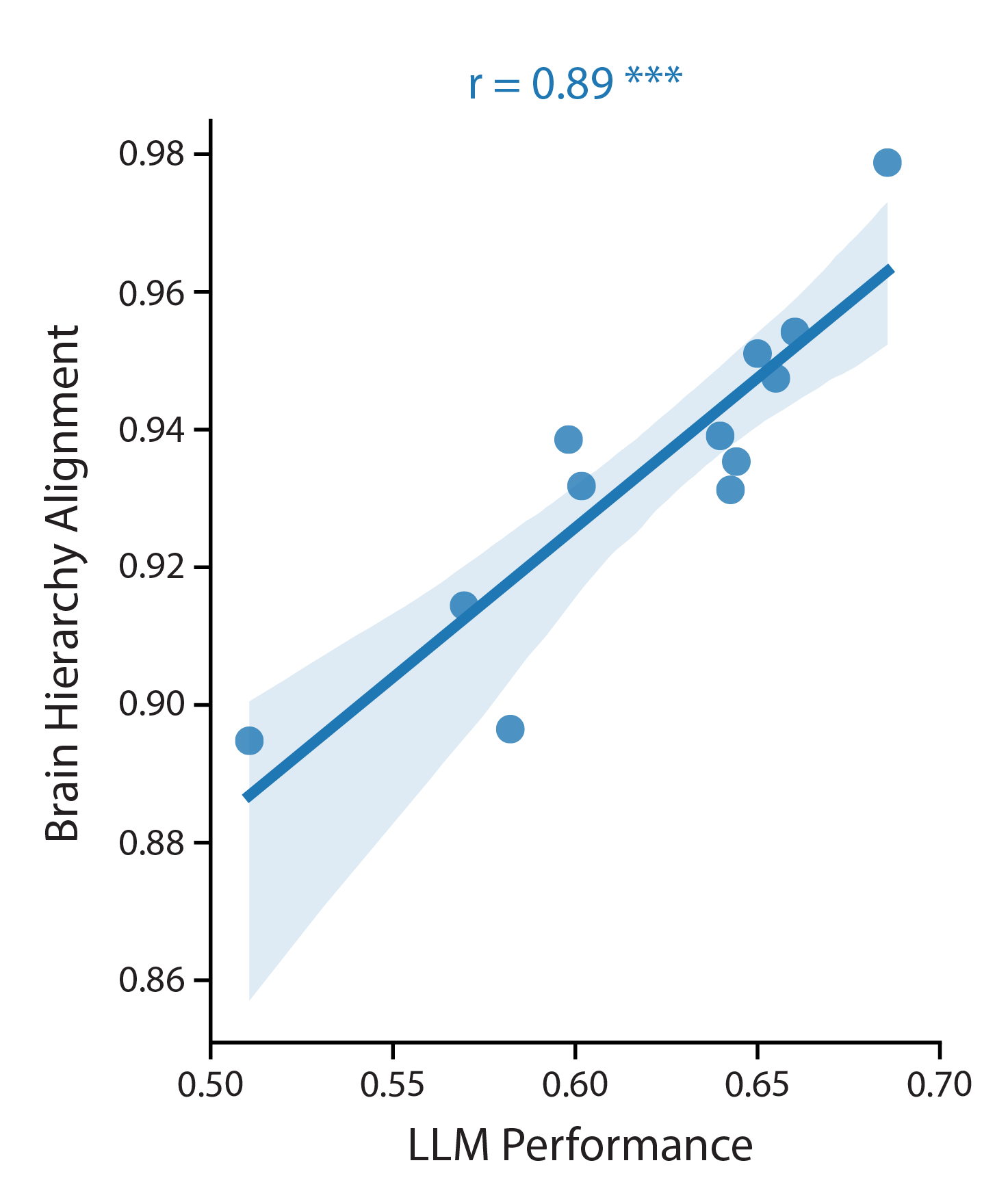}}
  \caption{Hierarchy alignment by model when using electrode lag instead of distance to estimate neural hierarchy. We used the electrode lag, instead of distance from primary auditory cortex, to bin electrodes into a hierarchy with a bin-width of 40ms. We estimated electrode lag using the peak of a 1D temporal receptive field fitted for each electrode to predict its response from the acoustic envelope of the stimulus sound. We then performed the same analysis as shown in Fig. \protect\ref{fig:3}, reproducing Fig. \protect\ref{fig:3}B with new brain hierarchy alignment for each model. These alignment values are similarly significantly correlated with LLM performance (Pearson $r=0.89, p=0.0001$).}
  \label{fig:s3}
\end{figure}

\begin{figure}[ht]
  \centering
  {\includegraphics[width=0.95\linewidth]{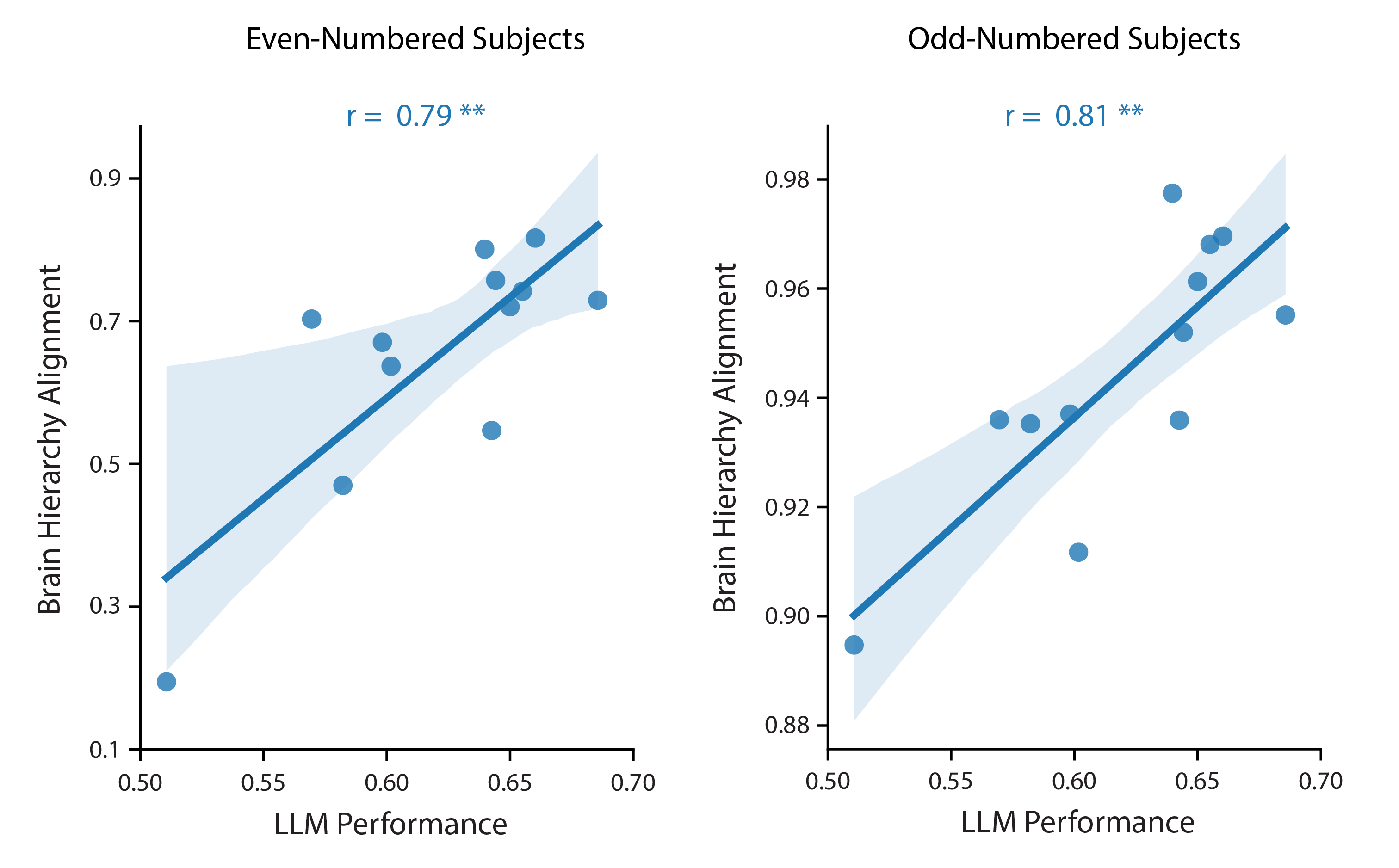}}
  \caption{Hierarchy alignment patterns hold for partial subject groupings. Splitting the electrodes based on whether they came from even- or odd-numbered subjects, we performed the same analyses as in Fig. \protect\ref{fig:3}B. Both subject groups show that brain hierarchy alignment is significantly correlated with LLM performance (Pearson correlations in figure, even $p=0.0022$, odd $p=0.0013$) demonstrating that this effect is not the result of a single outlier subject.}
  \label{fig:s4}
\end{figure}

\end{document}